\documentclass[final]{cvpr}

\usepackage{times}
\usepackage{epsfig}
\usepackage{graphicx}
\usepackage{amsmath}
\usepackage{amssymb}
\usepackage{textcomp}
\usepackage{xcolor}
\usepackage{tipa}
\usepackage{multirow}
\usepackage{enumitem}
\makeatletter
\newcommand{\printfnsymbol}[1]{%
\textsuperscript{\@fnsymbol{#1}}%
}

\usepackage[pagebackref=true,breaklinks=true,colorlinks,bookmarks=false]{hyperref}

\def\cvprPaperID{1835} 
\def\confYear{CVPR 2021}

\begin{document}

\title{Relation-aware Instance Refinement for Weakly Supervised Visual Grounding}

\author{Yongfei Liu\textsuperscript{\rm 1,5,6}\thanks{Both authors contributed equally. 
This work was done when Yongfei Liu was a research intern at Tencent AI Lab, and Bo Wan was a master student in ShanghaiTech University. This work was supported by Shanghai NSF Grant (No. 18ZR1425100).} 
\quad Bo Wan\textsuperscript{\rm 1,2}\printfnsymbol{1} \quad Lin Ma\textsuperscript{\rm 3} 
\quad Xuming He\textsuperscript{\rm 1,4}\\
\textsuperscript{\rm 1}School of Information Science and Technology, ShanghaiTech University \\
\textsuperscript{\rm 2}Department of Electrical Engineering (ESAT), KU Leuven \\
\textsuperscript{\rm 3}Meituan \quad  \textsuperscript{\rm 4}Shanghai Engineering Research Center of Intelligent Vision and Imaging\\
\textsuperscript{\rm 5}Shanghai Institute of Microsystem and Information Technology,
Chinese Academy of Sciences\\
\textsuperscript{\rm 6}University of Chinese Academy of Sciences\\

\{liuyf3, wanbo, hexm\}@shanghaitech.edu.cn \quad forest.linma@gmail.com

}

\maketitle
\begin{abstract}
   \vspace{-2mm}
   Visual grounding, which aims to build a correspondence between visual objects and their language
   entities, plays a key role in cross-modal scene understanding. One promising and scalable strategy for learning visual grounding is to utilize weak supervision from only image-caption pairs. Previous methods 
   typically rely on matching query phrases directly to a precomputed, fixed object candidate pool, which leads to inaccurate localization and ambiguous matching due to lack of semantic relation constraints. 
   In our paper, we propose a novel context-aware weakly-supervised learning method that incorporates coarse-to-fine object refinement and entity relation modeling into a two-stage deep network, capable of producing more accurate object representation and matching. To effectively train our network, we introduce a self-taught regression loss for the proposal locations and a classification loss based on parsed entity relations. 
   Extensive experiments on two public benchmarks Flickr30K Entities and ReferItGame demonstrate the efficacy of our weakly 
   grounding framework. The results show that we outperform the previous methods by a considerable margin, 
   achieving 59.27\% top-1 accuracy in Flickr30K Entities and 37.68\% in the ReferItGame dataset respectively\footnote{Code is available at https://github.com/youngfly11/ReIR-WeaklyGrounding.pytorch.git}.
\end{abstract}

\vspace{-5mm}
\section{Introduction}


Cross-modal understanding of visual scene and natural language description   
plays a crucial role in bridging human and machine intelligence, and has attracted much interest from AI community~\cite{kafle2019challenges}. 
Towards this goal, one core problem is to establish instance-level correspondence between visual regions and its related language 
entities, which is commonly referred to as visual grounding~\cite{karpathy2014deep}. Such correspondence serves as a fundamental 
building-block for many vision-language tasks, such as image captioning~\cite{feng2019unsupervised, wang2019controllable}, visual 
question answering~\cite{zellers2019recognition,mun2018learning}, visual navigation~\cite{wang2019reinforced,zhu2020vision} and 
visual dialog\cite{kottur2018visual,guo2020iterative}.

Much progress has been made recently in learning visual grounding with strong supervision~\cite{liu2020LCMCG,yang2020graph}, which requires costly annotations on region-phrase correspondence. 
A more scalable modeling strategy is to learn from only image-caption pairs, 
namely \textit{weakly-supervised visual grounding}~\cite{rohrbach2016grounding,yeh2018unsupervised,chen2018knowledge,gupta2020contrastive,liu2019adaptive}. 
Nevertheless, learning from such weak supervision is particularly challenging 
mainly due to the severe ambiguity 
in visual object location and in correspondence between diverse noun phrases and object entities during cross-modal learning. 
    

\begin{figure}
	\centering
	\includegraphics[width=0.9\linewidth]{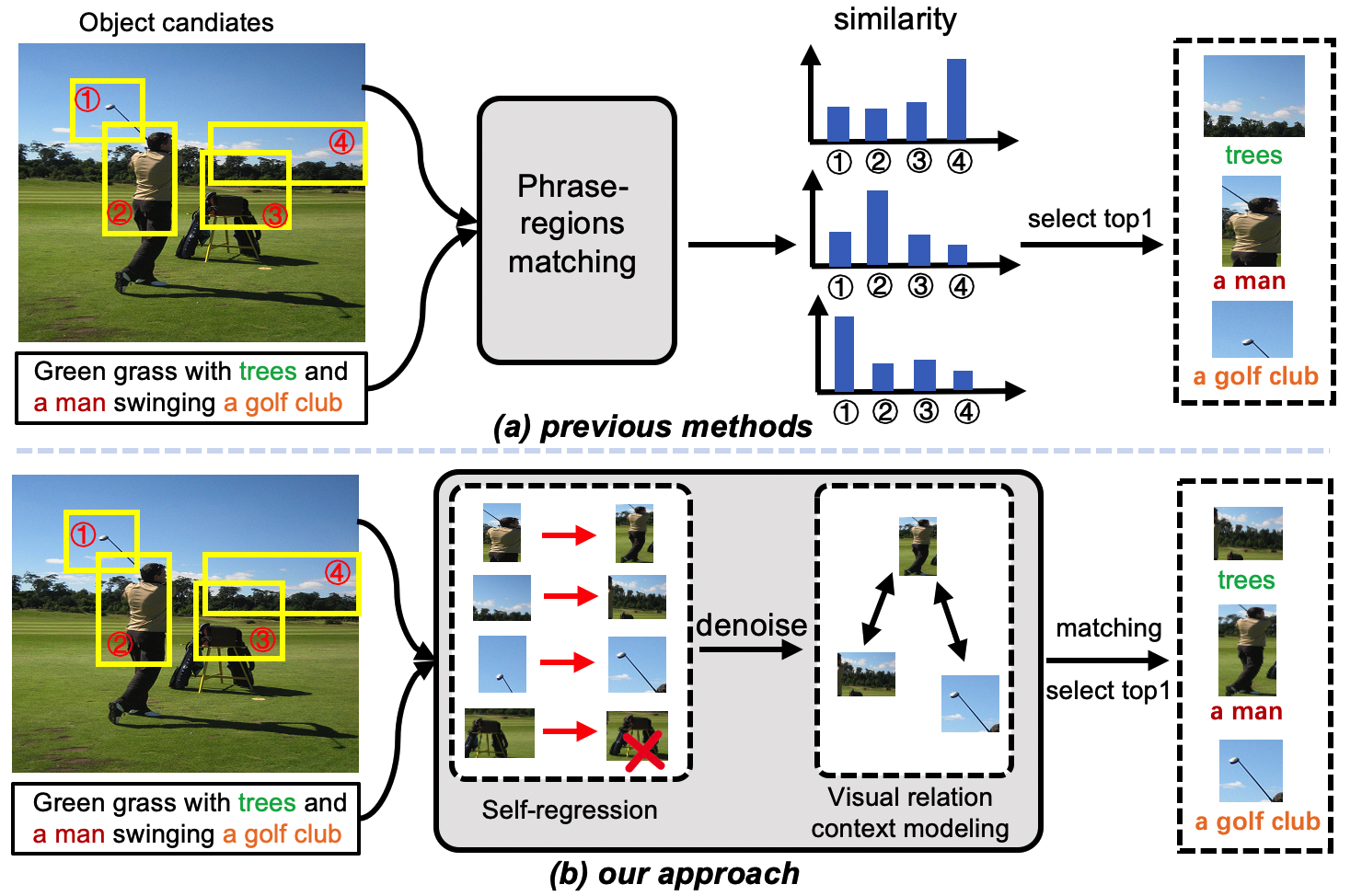}
	\caption{\small{Comparison of visual entities representation with existing weakly-supervised grounding models.
	(a) Previous methods directly match between noun phrases and a precomputed, fixed object proposals.
	(b) Our approach is capable of refining the initial object proposals and enriching their representation with visual relation context cues.}}\vspace{-3mm}
	\label{fig:ads}
\end{figure}
Most existing approaches tackle those challenges 
via the Multiple Instance 
Learning~(MIL) framework~\cite{karpathy2015deep} 
using object candidates generated by a pre-trained object detector~\cite{rohrbach2016grounding,datta2019align2ground,chen2018knowledge}. 
Despite their promising results, these learning pipelines often suffer from the visual and matching ambiguity from several aspects. 
First, they usually rely on a precomputed object proposal set that contain many distractor or background regions, making it 
difficult to infer positive matches for learning. 
In addition, these proposals are typically kept fixed during learning, which leads to inaccurate localization bounded 
by the external detectors (cf. Fig.~\ref{fig:ads}{(a)}). Furthermore, these methods often represent noun phrase or visual object context in an implicit 
manner, using attention-based feature aggregation or encoding predicate triples~\cite{liu2019adaptive,liu2019knowledge}. 
Such representations are limited in capturing rich semantic constraints from relations in an image-sentence pair, 
resulting in cross-modal matching ambiguity in both learning and prediction.

To address the afore-mentioned limitations, we propose a flexible and context-aware object representation for 
weakly-supervised visual grounding in this work. Unlike previous work, our representation is capable of refining the spatial locations of object proposals using a self-taught mechanism, and incorporates a relation-aware context model by exploiting the language prior (cf. Fig.~\ref{fig:ads}{(b)}). 
Such enriched representation alleviates the impact from the inaccurate object detection and the cross-modal matching ambiguity. 
To achieve this, we develop a coarse-to-fine matching strategy modeled as a two-stage deep network. The first stage of our model consists of a backbone and a coarse-level matching network for proposal generation and refinement, while the second stage builds a visual object graph network and a fine-level matching network for context modeling and final matching prediction.  

Specifically, given a pair of image and language description,
we first use the backbone network to generate a set of object proposals
with their visual features and compute the language
embedding for the noun phrases. Then the coarse-level matching network selects a small set of relevant proposals for each
phrase and refines their spatial locations. For the second stage, we construct the visual object graph network on the 
refined proposals by exploiting parsed language
structure, which enriches object features with their relations and context. Based on the context-aware representation, 
the fine-level matching network finally predicts instance-level correspondence between phrases and object proposals,
as well as further refined object locations. 

To train our deep network in a weak supervision setting, we introduce a novel multi-task loss function to exploit both the model prediction and linguistic relation cues. In particular,  
we first devise a self-taught regression loss for the proposal location refinement, which employs highly confident proposal predictions as pseudo groundtruth for their neighboring proposals. 
Moreover, we develop a classification loss on visual relation types based on the output of an external language parser. This enables us to generate effective supervision from the noisy language parsing results for learning better entity representations. 

We conduct extensive experiments on two public benchmarks:
Flickr30K Entities~\cite{plummer2015flickr30k} and
ReferItGame~\cite{kazemzadeh2014referitgame}.
The experiment results show that our method outperforms the prior
state-of-the-art with a considerable margin. To validate the effectiveness
of each model component, we also provide the detailed ablative study on
Flickr30K Entities dataset.
The main contributions of our work are three-folds:
\begin{itemize}[topsep=2pt,itemsep=0pt]
	\item We adopt a coarse-to-fine strategy to refine object proposals and alleviate semantic ambiguities by enriching visual feature with relationship constraints.
	\item We propose a self-taught regression loss to supervise object proposal refinement, and introduce an additional visual
	relation loss that helps learn a context-aware object representation.
	\item Our method achieves new state-of-the-art performance on Flickr30K Entities and ReferItGame benchmarks.
\end{itemize}

\section{Related Work}
\paragraph{Visual Grounding}
Visual grounding~\cite{plummer2018conditional,yu2018rethinking,liu2020LCMCG,SeqGROUND,yu2018mattnet,chen2017query} aims to learn region-phrase correspondence 
\textbf{with bounding box annotation} for each phrase during training stage. 
In recent years, the deep network is widely used in this task and achieves remarkable success.
Plummer et al.~\cite{plummer2018conditional} devised a single end-to-end network to learn multiple text-conditioned embedding for grounding 
and DDPN~\cite{yu2018rethinking} proposed to generate a group of high-quality proposals with a diversified and discriminate network. 
However, they ignored the semantic context cues and relation constraints in both vision and language. 
To address this problem, Nagaraja et al.~\cite{nagaraja2016modeling} explored to utilize LSTM to encode visual and linguistic context for referring expression, and SeqGROUND~\cite{SeqGROUND} adopts chain-structure LSTMs to encode context in cross-domain with a history stack for visual grounding. 
Besides, Wang et al.~\cite{wang2019neighbor} took a self-attention mechanism to capture their context in a sentence and build a directed graph over neighbor objects to exploit the visual relations, 
and Liu et al.~\cite{liu2020LCMCG} aimed to build a cross-modal graph network under the guidance of language structure to learn global context representation for both phrases and visual objects.
Although these methods demonstrate superior performance on visual grounding, they highly rely on the strong supervision that is
too expensive to obtain in most scenarios. Thus the main focus of this work is on learning cross-modal matching in a weak supervision setting.

\vspace{-3mm}
\paragraph{Weakly-Supervised Visual Grounding}
Different from supervised visual grounding, a weakly-supervised setting aims to learn the fine-grained region-phrase correspondence \textbf{with only image-sentence association}.
Most recent works~\cite{rohrbach2016grounding,yeh2018unsupervised,chen2018knowledge,datta2019align2ground,gupta2020contrastive,wang2020improving,liu2019adaptive, wang2020maf} take a hypothesis-and-matching strategy for
the weakly-supervised visual grounding task, in which they first generate a set of region proposals from an image with an external object detector, and then match between query phrases and those regions. WPT~\cite{wang2019phrase} directly computed the cross-modal similarity between noun phrases and detected multi-level visual concepts from amounts of object detectors.
GroundR~\cite{rohrbach2016grounding} built correspondences by reconstructing phrases with an attention mechanism on visual features.
To explore more powerful supervision, KAC-Net~\cite{chen2018knowledge} took a similar formulation but exploited visual consistency and knowledge
from object categories, and ~\cite{datta2019align2ground} adopted a ranking-loss to minimize the distances between associated image-caption and maximize the distance between irrelevant pairs.
Besides, ~\cite{gupta2020contrastive,wang2020improving} introduced a contrastive loss to distillate knowledge from external 
language~\cite{devlin2018bert} and visual models~\cite{joseph2016yolo}. 

Although these methods discovered various type of supervision for weakly-supervised visual grounding, they suffered from limited objects recall and all the methods above fail to refine object regions 
due to the lack of location supervision.
MATN~\cite{zhao2018weakly} solved this problem by introducing a transformation network to search target phrase location over the entire image directly, 
and such locations were regularized by the precomputed proposals.
Only a few work took into account context cues to eliminate semantic ambiguities in a weakly-supervised setting: ARN~\cite{liu2019adaptive} extracted the linguistic and visual cues on entity, location and context levels separately that enforced multi-level cross-modal consistency. KPRN~\cite{liu2019knowledge} further exploited linguistic context and required subject \& object matching simultaneously. 
Our focus is to exploit context-aware instance refinement weakly-supervised learning strategy for both limitations.

\begin{figure*}
	\centering
	\includegraphics[width=\linewidth]{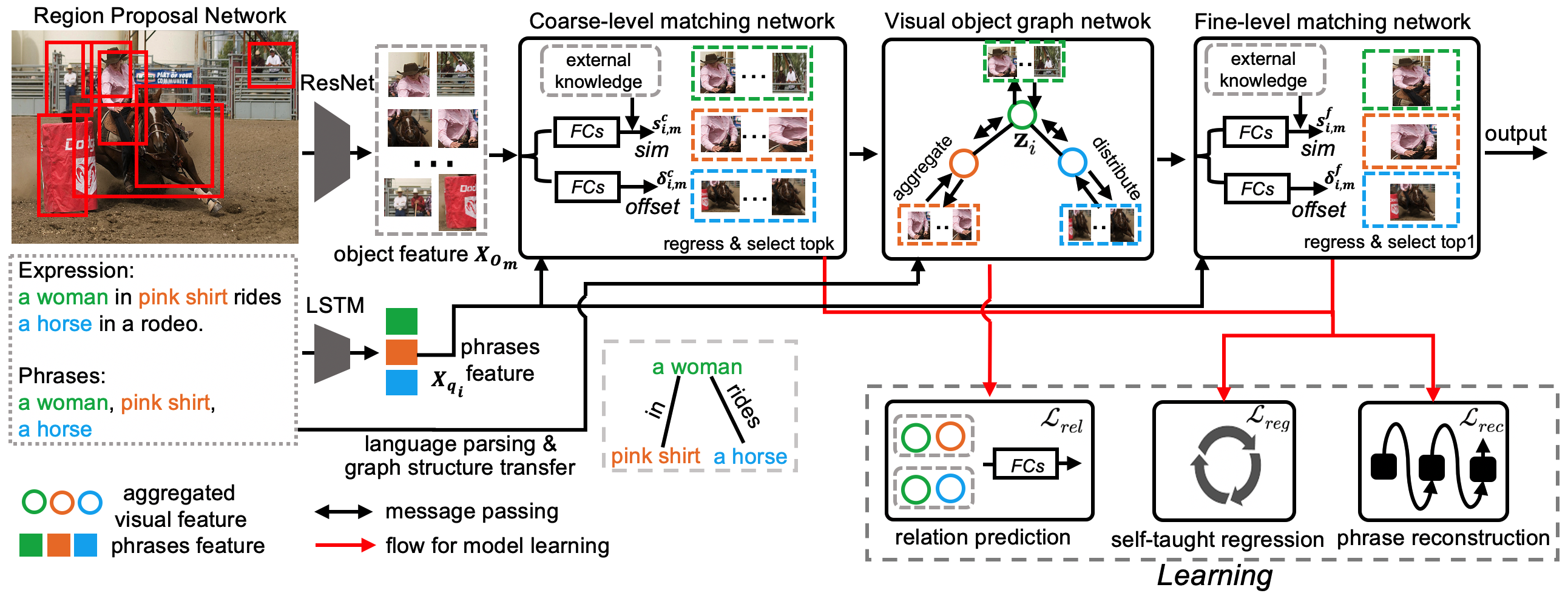}
	\caption{\small{\textbf{Model Overview:} There are four modules in our network, the \textbf{Backbone Network} prepares 
    basic phrase and visual features; the \textbf{Coarse-level Matching Network} selects a small set of objects for each phrase 
    and refines their spatial locations; the \textbf{Visual Object Graph Network} enriches the object feature with their context
    and relations by exploiting language structure, finally the \textbf{Fine-level Matching Network} predicts
    instance-level correspondences and refines their locations further based on the context-aware visual representation.
    Three main losses are demonstrated to supervise the whole network, like relation classification loss $\mathcal{L}_{rel}$, 
    self-taught regression loss $\mathcal{L}_{reg}$ and phrase reconstruction loss $\mathcal{L}_{rec}$.}} \vspace{-3mm}
	\label{fig:overview}
\end{figure*}

\section{Method}
\subsection{Problem Setting and Overview}

The task of weakly-supervised phrase grounding aims to localize the noun phrases of a language description in an associate image, 
while the correspondences between the noun phrases and image regions are not available for training.
Formally, we aim to learn a visual grounding model $\mathcal{M}$, which takes an input image $I$ and a description $D$ with a group of noun phrases $\mathcal{Q}=\{q_i\}_{i=1}^N$ and predicts the corresponding locations $\mathcal{B}=\{b_i\}_{i=1}^N$
for the query phrases, i.e.,
$\mathcal{B} = \mathcal{M}(I, D, \mathcal{Q})$. 
For the weakly-supervised learning setting, we are only provided a training dataset $\mathcal{X}=\{(I^{(l)}, D^{(l)}, \mathcal{Q}^{(l)})\}_{l=1}^L$ of size $L$, where the corresponding locations $\mathcal{B}^{(l)}$ of phrases $\mathcal{Q}^{(l)}$  are unobserved. 




In this work, we adopt a typical grounding strategy that first uses an external object detector (e.g., \cite{ren2015faster}) to generate a group of visual object proposals $\mathcal{O}$ from $I$, which are then matched to the phrases~\cite{rohrbach2016grounding,yeh2018unsupervised,chen2018knowledge}. Learning a cross-modal matching with weak supervision, however, is particularly challenging in such a generate-and-match framework due to visual ambiguity caused by inaccurate object detection and the lack of instance-level region-phrase correspondence.   

To address those challenges, we propose to learn a flexible context-aware entity representation based on the language prior
and a coarse-to-fine matching process, which enables us to mitigate the impact of the matching and localization ambiguity. 
We instantiate our strategy with a two-stage deep network, as illustrated in Fig.~\ref{fig:overview}. Specifically, the first-stage network extracts the visual and textual features from inputs, and perform a coarse-level matching in which we estimate the similarity scores between each $q_i$ and $\mathcal{O}$ and a refinement of object proposal locations. In the second stage, we select a small group of relevant proposals from $\mathcal{O}$ for each phrase $q_i$ according to the similarity scores, and build a visual object graph network by exploiting parsed language structure. Our second-stage network performs message passing to enhance the visual representation with contextual cues, and finally predicts a fine-level similarity score for each object-phrase pair as well as further refinement of object locations. 

To train our model, we develop a joint learning strategy with a multi-task loss, which additionally incorporates a self-taught regression loss to refine the object locations and a language-induced relation classification loss to enforce a relational constraint on the entity matching. Below we first introduce the details of our model architecture in Sec.~\ref{sec:model} followed by our design of loss functions in Sec.~\ref{sec:learning}.

\subsection{Model Architecture}\label{sec:model}

We now introduce our two-stage network which 
consists of four sub-modules and can be divided into two stages. The first stage of our network includes a \textit{Backbone Network} to extract visual and linguistic features (Sec.~\ref{sec:backbone}), and a \textit{Coarse-level Matching Network} to refine object locations and select a subset of relevant proposals for each phrase (Sec.~\ref{sec_coarsenet}). For the second-stage network, we build a \textit{Visual Object Graph Network} to capture visual context cues by message passing (Sec.~\ref{sec:vogn}), and a \textit{Fine-level Matching Network} to compute the final matching and object locations with context-aware features (Sec.~\ref{sec:finenet}).

\vspace{-3mm}
\subsubsection{Backbone Network}\label{sec:backbone}

Our first module is a backbone network consisting of a convolutional network for extracting visual features and a recurrent network for encoding language features. 

The convolutional network (e.g.~ResNet~\cite{he2016resnet}) takes the image $I$ as input and computes a feature map $\mathbf{\Gamma}$. An external object detector (e.g. Faster R-CNN~\cite{ren2015faster}) 
provides a set of object proposals $\mathcal{O}=\{\langle o_m, c_m \rangle\}_{m=1}^M$, where $o_m\in \mathbb{R}^4$ denotes object regions and $c_m\in \{1,2,...,C\}$ indicates 
object category. Similar to~\cite{liu2020LCMCG}, for each $o_m$, we use RoI-Align~\cite{he2017mask} and global average pooling to compute its conv-feature,
which is fused with its spatial feature and embeded to a vector 
$\mathbf{x}_{o_m}\in \mathbb{R}^{d}$, where $d$ is the feature dimensions.

For the language features, we compute an embedding of noun phrases $q_i \in \mathcal{Q}$ as follows. We first embed the words in description $D$ 
into a sequence of vectors $\mathbf{H}=\{\mathbf{h}_t\}_{t=1}^T$ via an encoding LSTM~\cite{hochSchm97}, where $T$ is the word length in $D$. The language feature $\mathbf{x}_{q_i}$ of each phrase $q_i$ is computed 
by taking average pooling on its word representations:
\begin{align}
\label{phrase}
\mathbf{H} = \emph{{\rm LSTM}}_{enc}(D),\quad
\mathbf{x}_{q_i} = \frac{1}{|q_i|}\sum_{t\in q_i} \mathbf{h}_t
\end{align}
where $|q_i|$ indicates the phrase length in words, and the embeddings $\mathbf{h}_t,\mathbf{x}_{q_i}\in\mathbb{R}^{d}$, which have the same dimensionality as the visual features. 

%
\vspace{-2mm}
\subsubsection{Coarse-level Matching Network}\label{sec_coarsenet}

Our second module performs a coarse-level matching between phrases and the initial object proposals, 
 aiming to select a small set of relevant proposals and refine their spatial locations.   
To this end, for each phrase-boxes pair $\{q_i, o_m\}$, we compute a similarity score  $\hat{s}_{i,m}^c$ and regression offsets 
$\delta_{i,m}^c \in \mathbb{R}^{4}$ according to the phrase feature $\mathbf{x}_{q_i}$ and object feature $\mathbf{x}_{o_m}$ as follows:
\begin{align}
\hat{s}_{i,m}^c = F_{cls}(\mathbf{x}_{q_i},\mathbf{x}_{o_m}),\quad
\delta_{i,m}^c = F_{reg}(\mathbf{x}_{q_i},\mathbf{x}_{o_m}) \label{equ:cls_reg}
\end{align}
where $F_{cls}$ and $F_{reg}$ are two fully-connected networks. 

Following \cite{chen2018knowledge}, we further utilize object categories as a semantic cue to compute an additional similarity score $s_{i,m}^a$ in the linguistic space:
\begin{align}
s_{i,m}^a & = \langle E_{ext}(q_i), E_{ext}(c_m) \rangle
\end{align}
where $\langle \cdot,\cdot\rangle$ indicates the inner product, and $E_{ext}$ represents an off-the-shelf 
language embedding (e.g., Skip-thoughts~\cite{kiros2015skip}). 
Finally, we fuse the above two similarity scores and compute an attention weight by taking the Softmax over all the proposals: 
\begin{align}
s_{i,m}^c  = \hat{s}_{i,m}^c \cdot s_{i,m}^a,  &\quad  \alpha_{i,m}^c = \underset{m\in [1:M]}{\emph{{\rm Softmax}}} (s_{i,m}^c)   \label{equ_atten}
\end{align}

To refine the proposal set, we apply the regression offsets $\boldsymbol{\delta}_i^c  = \{\delta_{i,m}^c\}_{m=1}^M$ and select the top $K (K\ll M)$ proposals for each phrase $q_i$ based on the similarity 
scores $\mathbf{s}_i^c = \{s_{i,m}^c\}_{m=1}^M$. This generates a set of refined proposals $\mathcal{V}_i = \{o_{i,k}\}_{k=1}^K$ for phrase $q_i$, and we denote the proposal set for all the noun phrases as $\mathcal{V}= \{\mathcal{V}_i\}_{i=1}^N$.


\vspace{-2mm}
\subsubsection{Visual Object Graph Network}\label{sec:vogn}
Given the refined proposal sets, we introduce a graph neural network, dubbed as the Visual Object Graph Network, to capture the visual context with the guidance of language structure. Specifically, we first extract a set of relation phrases from the description $D$ with an external language parser\footnote{1
https://github.com/vacancy/SceneGraphParser}~\cite{schuster2015generating, wang2018sgparser,liu2020LCMCG}. 
We then build a graph network with $N$ nodes and its $i$-th node, denoted by $\mathbf{z}_i$, encodes the visual feature for phrase $q_i$. Two nodes $\mathbf{z}_i$ and $\mathbf{z}_j$ are connected if a relation phrase exists between two phrases $q_i$ and $q_j$. 

We initialize the node feature $\mathbf{z}_i$ based on the phrase proposal set $\mathcal{V}_i$ and its attention scores $\{\alpha_{i,k}^c\}_{k=1}^K$, which represents an estimation of the visual feature for $q_i$:
\begin{align}
\mathbf{z}_{i} = \sum_{k=1}^{K}  \alpha_{i,k}^c \cdot \mathbf{x}_{o_{i,k}}
\end{align}
where $\mathbf{x}_{o_{i,k}}$ are the object proposal features, and $ \alpha_{i,k}^c $ are from Eq.~\ref{equ_atten}. Subsequently, our graph network refines the visual features $\{\mathbf{z}_{i}\}$ by a message passing step as below: 
\begin{align}
\mathbf{z}_{i}'&= \mathbf{z}_{i} + \sum_{j\in \mathcal{N}(i)} \omega_{i,j} F_M(\mathbf{z}_{j})\\
\omega_{i,j} &= \underset{j\in\mathcal{N}(i)}{\emph{{\rm Softmax}}} (F_M(\mathbf{z}_{i})^\intercal F_M(\mathbf{z}_{j}))
\label{equ:attention}
\end{align}
where $\mathbf{z}_{i}'$ denotes the updated visual features, $\mathcal{N}(i)$ is the neighborhood of node $i$, $F_M$ is 
a multi-layer network for computing messages, and $\omega_{i,j} $ is an attention weight between node $i$ and $j$.
Finally, we update each object proposal feature, denoted by $\{\mathbf{x}_{o_{i,k}}'\}$, with the visual features of its corresponding phrase $q_i$ as follows:
\begin{align}
\mathbf{x}_{o_{i,k}}' = \mathbf{x}_{o_{i,k}} + \alpha_{i,k}^c \cdot \mathbf{z}_{i}',
\end{align}
where the visual context in $\mathbf{z}_{i}'$ is distributed to $o_{i,k}$ with attention weighting.

\vspace{-2mm}
\subsubsection{Fine-level Matching Network}\label{sec:finenet}
Given the context-aware features, our final module performs a fine-level matching between phrases and the refined object proposal subsets. Similar to the coarse-level matching, we predict a similarity score $s_{i,k}^f$, an attention weight $\alpha_{i,k}^f $, and a phrase-specific regression 
offset $\delta_{i,k}^f \in \mathbb{R}^{4}$ for each phrase-proposal pair $\{q_i, o_{i,k}\}$ as below:
\begin{align}
	{s}_{i,k}^f = F_{cls}(\mathbf{x}_{q_i},&\mathbf{x}_{o_{i,k}}')\cdot s_{i,k}^a, {\rm{\ \ }} \alpha_{i,k}^f =  \underset{k\in[1:K]}{\emph{{\rm Softmax}}}(s_{i,k}^f) \\
	&\delta_{i,k}^f = F_{reg}(\mathbf{x}_{q_i},\mathbf{x}_{o_{i,k}}') \label{equ:cls_reg_s2} 
\end{align}
\vspace{-8mm}
\paragraph{Model Inference}
During the model inference, for each query $q_i$, we first compute an overall matching scores $\{ s_{i,k} \}_{k=1}^K$ for the candidate proposals in $\mathcal{V}_i$ by fusing the coarse-level and fine-level scores:
\vspace{-2mm}
\begin{align}
	s_{i,k} = s_{i,k}^c \cdot s_{i,k}^f,
\end{align}
followed by applying the estimated offset $\delta_{i,k}^f$ to their locations. Finally, we take the proposal $o_{i,k}^*$ with the maximum similarity score $s_{i,k}^*$ as the grounding result of phrase $q_i$.

\subsection{Learning with Weak Supervision}\label{sec:learning}

We now introduce our weakly supervised learning strategy for training the two-stage deep network. To this end, we develop a multi-task loss that incorporates two novel supervision signals from a partially trained model itself and a language prior on entity relations, respectively. 

Specifically, our overall loss function  $\mathcal{L}$ consists of four terms, including a reconstruction loss $\mathcal{L}_{rec}$ for noun phrases, a self-taught regression loss $\mathcal{L}_{reg}$ for refining object proposal locations, a relation classification loss $\mathcal{L}_{rel}$ for language-induced relation cues, and a ranking loss $\mathcal{L}_{rank}$ for image-caption pairs. Formally, this weakly-supervised learning loss can be written as follows,
\begin{align}
\mathcal{L} =  \mathcal{L}_{rec}
+ \lambda_1 \cdot \mathcal{L}_{reg} + 
\lambda_2 \cdot \mathcal{L}_{rel} + \lambda_3 \cdot \mathcal{L}_{rank}
\end{align}
where $\{\lambda_1, \lambda_2, \lambda_3\}$ are weight coefficients to balance the loss terms. In this work, we adopt a similar ranking loss as \cite{datta2019align2ground}, and leave its details to the Suppl. We will instead focus on the remaining three loss terms below, which are defined for each image-caption pair.  


\begin{figure}[t]
	\centering
	\includegraphics[width=0.65\linewidth]{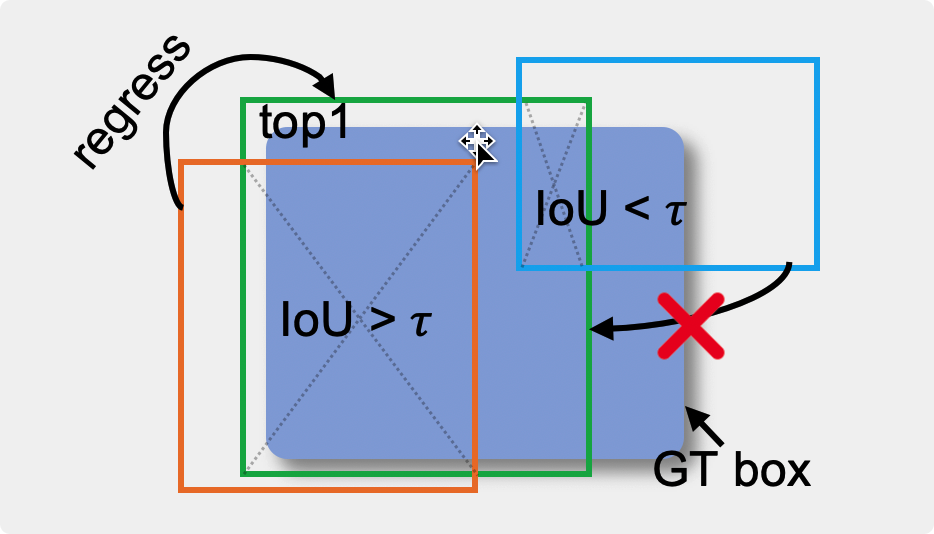}
	\caption{\small{Illustration of self-taught regression. The green box has the highest confident matching score. 
		The orange box overlapping with the green box with IoU $> \tau $ will be regressed toward the green box, while the blue box overlapping with the green box with IoU $< \tau $ will stay unchanged.
		The GT box region here is unobserved during training.}}\vspace{-3mm}
	\label{reg_ill}
\end{figure}

\vspace{-3mm}
\paragraph{{Phrase reconstruction loss $\mathcal{L}_{rec}$}} 

Given the noun phrases, we adopt a phrase reconstruction loss~\cite{chen2018knowledge,rohrbach2016grounding} at both stages of our deep network to provide model supervision. As those two reconstruction loss terms are similar,
we will use the coarse-level as an example below.

To apply the reconstruction loss, we first generate a visual representation $\mathbf{z}_{i}^c $ for phrase $q_i$. Specifically, we aggregate the object 
features $\{\mathbf{x}_{o_m}\}_{m=1}^M$ with the attention weights $ \{ \alpha_{i,m}^c \}_{m=1}^M$ in Eq.~\ref{equ_atten}, which can be written as
	$\mathbf{z}_{i}^c = \sum_{m=1}^{M}  \alpha_{i,m}^c \cdot \mathbf{x}_{o_m}$. 
We then reconstruct the noun phrase $q_i$ using the visual feature $\mathbf{z}_{i}^c$ based on a decoding LSTM. Concretely, we compute a sequence of word distribution $ \hat{\mathbf{y}}_i^c = \{\hat{\mathbf{y}}_{i,w}^c \}_{w=1}^{|q_i|} $ as below:
\begin{align}
\hat{\mathbf{y}}_i^c = \emph{{\rm LSTM}}_{dec}([\mathbf{z}_{i}^c, q_{i}])\label{equ:decode}
\end{align}
Similarly, we also predict $ \hat{\mathbf{y}}_i^f$ in the fine-level based on a context-aware feature for each phrase $q_i$ as $\mathbf{z}_{i}^f=\sum_{k=1}^K\alpha_{i,k}^f\cdot \mathbf{x}'_{o_{i,k}}$.
Both stages share the same parameters for the decoder $\emph{{\rm LSTM}}_{dec}$. For each phrase, we adopt the standard sequence log loss $L_{\log}$ 
and the overall reconstruction loss can be written as:
\begin{align}
\mathcal{L}_{rec} = \sum_{i=1}^N \left( L_{\log}(\hat{\mathbf{y}}_i^c, q_i) + L_{\log}(\hat{\mathbf{y}}_i^f, q_i) \right)
\end{align}
\vspace{-2mm}

\begin{table*}[t!]
	\centering
	\resizebox{0.8\textwidth}{!}{
		\begin{tabular}{ccccc|cc|cc}
			\hline
			\multirow{2}{*}{Methods} &\multirow{2}{*}{Backbone} &\multirow{2}{*}{Language} &\multirow{2}{*}{Proposals} &\multirow{2}{*}{Det Label}  & \multicolumn{2}{c|}{\textbf{Flickr30k}}& \multicolumn{2}{c}{\textbf{ReferItGame}}\\
			&    &					  &                                 &                            &  \textit{Acc\%} & \textit{PointIt\%} &  \textit{Acc\%} & \textit{PointIt\%}\\
			\hline
			SSS~\cite{javed2018learning}             & VGG18      & N/A   & -   & \texttimes                      & -              & 49.10              & -              & 49.90\\
			MultiGrounding~\cite{akbari2019multi}    & PNAS Net   & Elmo  & -   & \texttimes                  & -              & 69.19          & -              & 48.42 \\
			\hline
			GroundR~\cite{rohrbach2016grounding}     & VGG16      & LSTM  & SS & \texttimes & 28.94          & -              & 10.70          & - \\
			MATN~\cite{zhao2018weakly}               & VGG16      & LSTM  & SS & \texttimes & 33.10          & -               & 13.61         & -\\
			KAC Net~\cite{chen2018knowledge}         & VGG16      & LSTM  & SS & \checkmark & 38.71          & -              & 15.83          & -\\  
			\hline
			Align2Ground~\cite{datta2019align2ground}          & RN152      & LSTM  & FRCNN(VG)   & \texttimes & 11.20          & 71.00     &-&-                      \\
			UTG~\cite{yeh2018unsupervised}                     & N/A        & Glove & YOLOV2(COCO)     & \checkmark & 36.93          & -         & 20.91          & -      \\
			ARN~\cite{liu2019adaptive}                         & RN101      & LSTM  & MaskRCNN(COCO)   & \checkmark  &-&-                         & 26.19          & -              \\
			KAC Net*~\cite{chen2018knowledge}                  & RN101      & LSTM  & FasterRCNN(VG)   & \checkmark & 46.61          & 74.17     & 33.67          & 56.57  \\
			KPRN\cite{liu2019knowledge}                       & RN101      & LSTM  & MaskRCNN(COCO)   & \checkmark  &-&-                         & 33.87          & -      \\
			Contr. Learning~\cite{gupta2020contrastive}        & RN101      & Bert  & FasterRCNN(VG)   & \texttimes & 51.67          & 76.74     &-&-     				   \\
			Contr. Dist.~\cite{wang2020improving}              & RN101      & LSTM  & FasterRCNN(OI)   & \checkmark & 50.96          & -         & 27.59          & -      \\
			\hline
			ours                                              & RN101      & LSTM  & FasterRCNN(VG)   & \checkmark & \textbf{59.27} & \textbf{78.60} &\textbf{37.68} & \textbf{58.96}\\
			\hline
	\end{tabular}}
	\caption{\small{Comparison of phrases grounding accuracy on Flickr30K Entities and ReferitGame test sets. 
	* denotes the re-implementation using the same backbone and object proposals as ours. SS denotes the selective search and (VG),(COCO),(OI) denote
	the object detector pre-trained on Visual Genome, MSCOCO, OpenImage dataset.}}
	\label{results_all}
\end{table*}

\vspace{-5mm}
\paragraph{\textbf{Self-taught regression loss $\mathcal{L}_{reg}$}} 
As phrase locations are not annotated, we introduce a self-taught regression loss for training the location regressors. Particularly, we observe that the proposals with high matching scores often have an accurate localization after several rounds of training without $\mathcal{L}_{reg}$. This motivates us to use the confident proposals from partially-trained models to supervise the location refinement of their neighboring proposals, as shown in Fig.~\ref{reg_ill}. Concretely,
for phrase $q_i$, we denote $\boldsymbol{\delta}_i^{c*} =\{\delta_{i,m}^{c*}\} $ in which $\delta_{i,m}^{c*}$ is the offset between proposal $o_m$ and the most confident proposal
if their overlaps are larger than a threshold $\tau$, and otherwise the predicted $\delta_{i,m}^{c}$, which means they will stay unchanged. 
We then adopt the smooth-L1 loss $L_{sm}$~\cite{ren2015faster,he2017mask} for the offset regression:
\begin{align}
	\mathcal{L}_{reg} =  \sum_{i=1}^N \left( L_{sm}(\boldsymbol{\delta}_i^c,\boldsymbol{\delta}_i^{c*}) + L_{sm}(\boldsymbol{\delta}_i^f,\boldsymbol{\delta}_i^{f*})\right)
\end{align}

\vspace{-3mm} 
\paragraph{\textbf{Relation classification loss $\mathcal{L}_{rel}$}} 

We further introduce a pairwise loss on the context-aware features of phrases, which imposes a relational constraint for the context encoding and fine-level matching. 
Specifically, we first extract relation phrases on the entire dataset by the language parser as described in Sec.~\ref{sec:vogn},
and select $C_r$ most frequent relations to form a set of relationship categories $\mathcal{R}=\{0,1,2,\cdots,C_r\}$, where 0 indicates no relation.
Then we predict the relation type $\hat{\mathbf{y}}_{i,j}^r \in \mathbb{R}^{C_r}$ between a pair of $\{q_i, q_j\}$ 
according to their fine-level context-aware object features with a multi-layer network $F_{rel}$ as follows,
\begin{align}
\hat{\mathbf{y}}_{i,j}^r = F_{rel}(\mathbf{z}_{i}^f , \mathbf{z}_{j}^f ) .
\end{align}
Denote the relation labels of $\{q_i, q_j\}$ as $r_{ij}\in \mathcal{R}$, we use the cross entropy loss for the relation classification if $ r_{ij}>0 $:
\begin{align}
	\mathcal{L}_{rel} = \sum_{i,j}L_{ce}(\hat{\mathbf{y}}_{i,j}^r, r_{ij}).
\end{align}

\vspace{-1mm}
\section{Experiments}
In this section, we first depict the experimental setup and implementation details; then compare our model
with previous arts. Detailed ablation studies are also conducted to validate each components in our model. Finally, 
we demonstrate several qualitative results to show model efficacy.

\subsection{Datasets and evaluation metric}

\textbf{Flickr30K Entities}~\cite{plummer2015flickr30k} contains 29783 images for training, 1000 images for validation and 1000 images for test. Each
image is associated 5 captions. For \textbf{ReferItGame}~\cite{kazemzadeh2014referitgame} dataset, there are around 20k images and 130k query phrases.
Each object is referred by 1-3 query phrase and we follow the standard dataset split of~\cite{rohrbach2016grounding}. It is worthed noting that we ignore
the box annotations for the noun phrases on both datasets during the training stage.

\textbf{Evaluation Metric:} We consider a noun phrase grounded correctly when its predicted box has at least 0.5 IoU with its ground-truth location. 
The grounding accuracy \textit{Acc} (i.e., Recall@1) is the fraction of correctly grounded noun phrases. We also report the point game metric \textit{PointIt} for a clear comparison with 
previous methods~\cite{javed2018learning,akbari2019multi,datta2019align2ground}. Following \cite{datta2019align2ground}, we define a hit if the center of the predicted bounding box lies in anywhere inside the 
ground-truth region, and \textit{PointIt} is the percentage of these hits.

\subsection{Implementation Details}
We generate an initial set of $M$=50 object proposals with an external RPN~\cite{ren2015faster} 
pre-trained on Visual Genome~\cite{Ranjay2017vg} dataset, and predict their object categories with the classification head of Faster-RCNN~\cite{ren2015faster}.
Then we apply RoI-Align~\cite{he2017mask} to extract the object visual representation on feature map $\mathbf{\Gamma}$, which is the output of C4 block in ResNet-101 with channel dimension $d$=2048. 
In coarse-level matching network, we set $K$=5 to filter out most irrelevant proposals for each noun phrase. In addition, we select $C_r$=88 relations, of which frequency are greater
than 100.

For model learning, we train the entire network with SGD optimizer with an initial learning rate of 1e-3 and weight decay of 5e-4. The training iterations are up to 80k and the batch size of each is 40.
We decay the learning rate by 10 times in 32k and 40k respectively. The hyper-parameters $(\lambda_1, \lambda_2, \lambda_3)$ are set as $(0.1, 1, 1)$ in loss function.
The threshold $\tau$=0.6 in self-taught regression and $\mathcal{L}_{reg}$ is applied for training after 7.5k iterations. All optimal hyper-parameters are selected by conducting a grid search on validation set and applied to test set directly once fixed.
More details on ReferItGame are described in Suppl.

\subsection{Quantitative Results}
We compare our model with several previous works in terms of \textit{Acc} and \textit{PointIt} on both Flickr30K 
Entities~\cite{plummer2015flickr30k} and ReferItGame~\cite{kazemzadeh2014referitgame} datasets.

\vspace{-3mm}
\paragraph{Flickr30K Entities:}
As shown in Tab.~\ref{results_all}, our approach outperforms the prior methods by a considerable margin in both evaluation metric, achieving 59.27\%
on \textit{Acc} and 78.60\% on \textit{PointIt}. Compared with reconstruction-based methods, we can outperform KAC Net*~\cite{chen2018knowledge} by 12.66\% on \textit{Acc} 
and 4.43\% on \textit{PointIt}, which demonstrates that our carefully-designed regression loss and visual object graph can solve 
the spatial and semantic ambuigities simultaneously. When compared with recently proposed contrastive learning based methods~\cite{wang2020improving,gupta2020contrastive},
we can still improve the performance by 7.60\% on \textit{Acc} and 1.96\% on \textit{PointIt}, although Contrastive Learning~\cite{gupta2020contrastive} takes more powerful
Bert as their language model.\footnote{Comparison with concurrent work~\cite{wang2020maf} which is trained with additional supervision from object attributes refers to Suppl.} In addition, we demonstrate more detailed performance comparisons of per coarse class in Suppl. 


\begin{table}[t]
	\centering
	\resizebox{0.42\textwidth}{!}{
		\begin{tabular}{ccccc}
			\hline
			Methods  & TSD  & STR               & VOGN\&RC             & \textit{Acc\%} \\
			\hline
			baseline & -          & -                        & -                     & 48.18  \\
			& \checkmark & -                        & -                     & 50.80  \\
			& \checkmark & \checkmark(w/o $\mathbf{x}_{q_{i}}$) & -                     & 54.05  \\
			& \checkmark & \checkmark               & -                     & 56.88  \\
			& \checkmark & -                        & \checkmark            & 52.48  \\
			& \checkmark & \checkmark               & \checkmark(w/o attention) & 55.60  \\
			ours     & \checkmark & \checkmark               & \checkmark            & 58.30  \\
			
			\hline
	\end{tabular}}
	\caption{\small{Ablation Study on Flickr30K Entities val set.}}
	\label{flickr-aba}
\end{table}

\vspace{-3mm}
\paragraph{ReferItGame:}
In ReferItGame dataset, we achieve 37.68\% on \textit{Acc} and 58.96\% on \textit{PointIt}.
Our method outperform KPRN~\cite{liu2019knowledge} by 3.81\%
on \textit{Acc} and KAC Net*~\cite{chen2018knowledge} by 3.39\% on \textit{PointIt}, 
which further validates the effectiveness of our method.



\begin{figure}
	\center
	\includegraphics[width=0.8\linewidth]{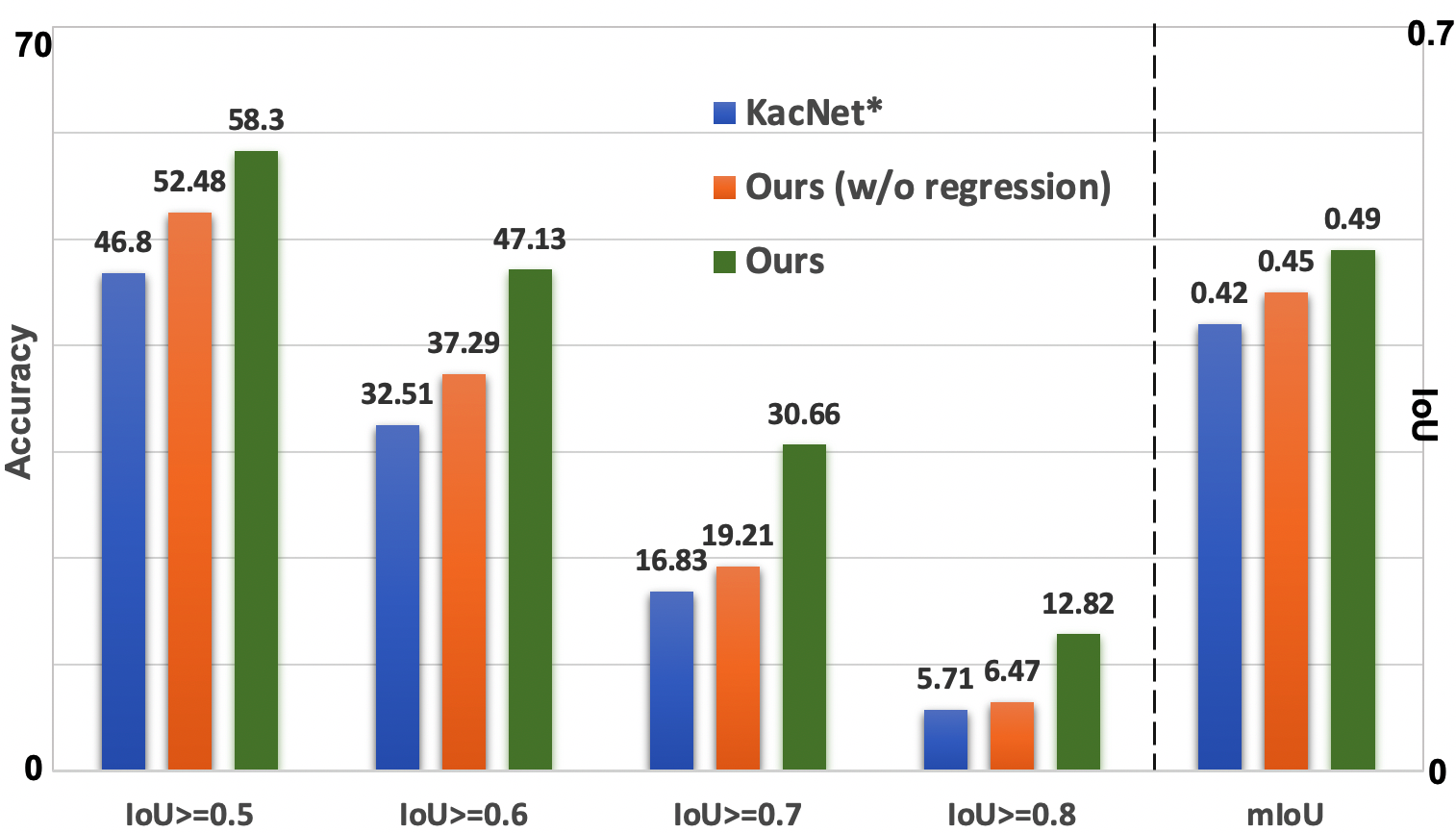}
	\caption{\small{Comparison of grounding accuracy in different IoU threshold and overall mean IoU on Flickr30K Entities val set.}}
	\label{fig:IOU_vis}
\end{figure}

\begin{figure*}[t!]
	\centering
	\includegraphics[width=0.85\linewidth]{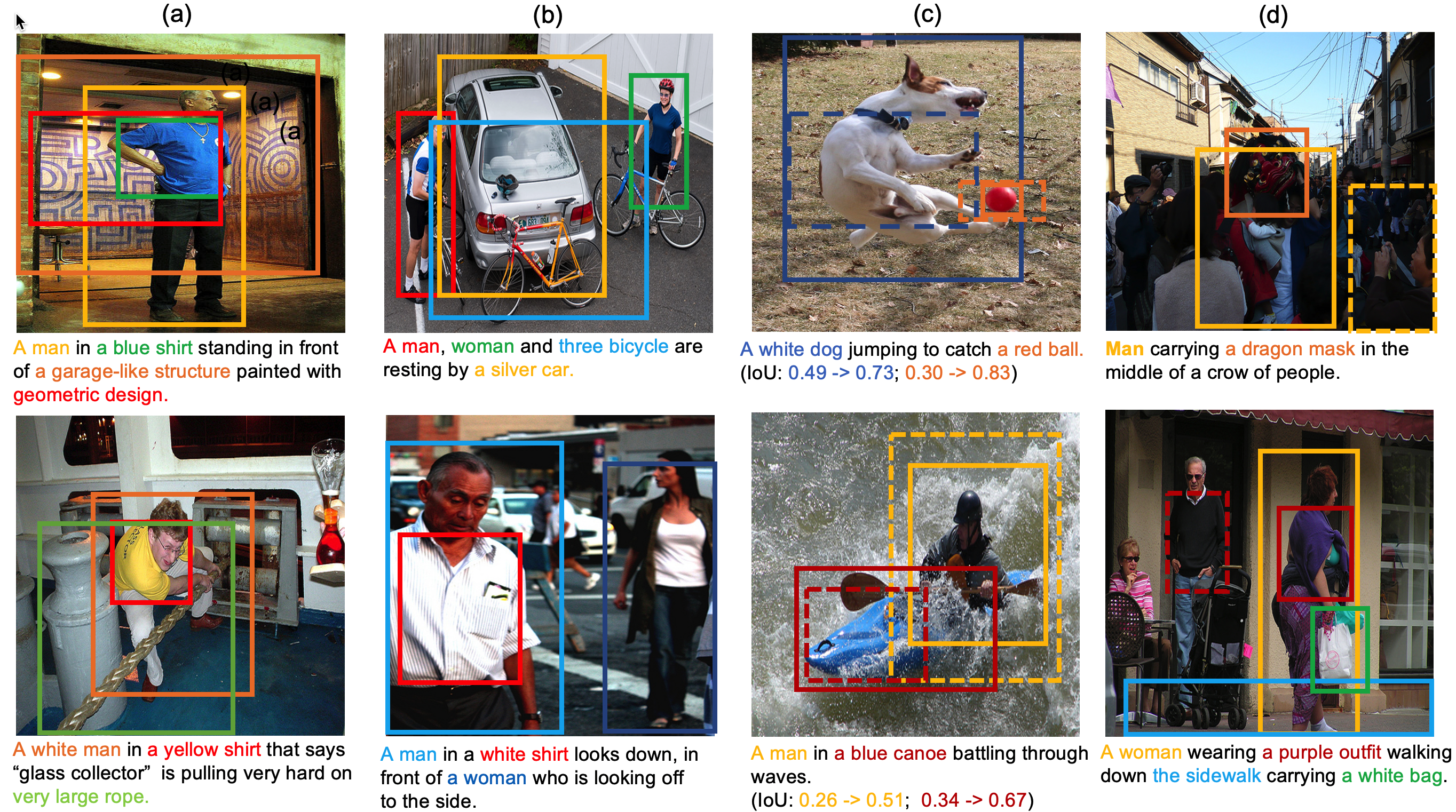}
	\caption{{\small Visualization of weakly grounding results on Flickr30K validation set. The colored boxes in image
			correspond to the noun phrases with the same color in sentence. \textbf{(a)} demonstrates grounding results when the input 
			sentences are complex, while \textbf{(b)} shows results when visual scene is complex. \textbf{(c)} shows the effects of 
			self-taught regression and \textbf{(d)} illustrates the results with the help of context cues. We denote that the dashed boxes in 
			\textbf{(c)} are initial proposals from external detectors, the solid boxes are our predictions after regression. In \textbf{(d)}, 
			the dashed boxes are the predictions from our model without visual object graph and relation constraints.}}
	\label{fig:demo}
\end{figure*}

\subsection{Ablation Study}
In this section, we conduct extensive ablation studies on Flickr30K Entities 
validation set to show effectiveness of each component in our method (Tab.~\ref{flickr-aba}).


\vspace{-3mm}
\paragraph{Baseline:} We take the direrctly matching strategy with only Backbone Network and Coarse-level Matching Network (w/o box regression)
as our baseline model, which is only supervised by the phase reconstruction loss $\mathcal{L}_{rec}$ and
and the ranking loss $\mathcal{L}_{rank}$.

\vspace{-3mm}
\paragraph{Two-stage denoising~(TSD):} As shown in Tab.~\ref{flickr-aba}, our two-stage denoising strategy can bring the performance gain of 2.64\% on \textit{Acc} compared with the baseline model. 
Because such strategy helps to filter out most of background distractors and irrelevant objects and thus alleviate difficulties in establishing cross-modal correspondence.

\vspace{-3mm}
\paragraph{Self-taught regression~(STR):} Different from the previous work~\cite{chen2018knowledge,liu2019knowledge,gupta2020contrastive,datta2019align2ground,liu2019adaptive}
whose performance is directly restricted to the quality of generated object proposals,
we improve the grounding accuracy from 50.8\% to 56.88\% by refining the location of object proposals and thus reduce spatial ambiguities under the supervision of self-taught box regression. 
In addition, we explore to remove noun phrase feature $\mathbf{x}_{q_i}$
in Eq.~\ref{equ:cls_reg} \& \ref{equ:cls_reg_s2} when estimating the proposal offsets based on visual features only, and find 
\textit{Acc} drop from 56.88\% to 54.05\%, which suggests the language feature
provides semantic-aware guidance for box regression.

To validate its effectiveness further, we remove such self-taught regression loss on the final model and observe a steep accuracy decrease in
different IoU threshold across from $0.5$ to $0.8$, as shown in Fig.~\ref{fig:IOU_vis} (left); and a mean IoU drop shown in Fig.~\ref{fig:IOU_vis} (right).

\vspace{-3mm}
\paragraph{Visual object graph network~(VOGN) \& Relation constrain~(RC):} Visual object graph network enriches each visual representation with their context cues 
and relation constrain provides additional supervision to learn such representation, as a result it suppresses semantic ambiguity and improve the accuracy from 56.88\% to 58.30\%.
It worth noting that VOGN incorporates with RC to work as a whole in our model, and we find using any separate component
will result in a limited contribution to the final results. More details refer to the Suppl.

To further explore the graph structure, we replace
the attention weights $\omega_{ij}$ in Eq.~\ref{equ:attention} with non-parametric uniform values by averaging the number of edges. 
We observe a significant performance drop from 58.30\% to 55.60\%, which suggests that it is non-trivial
to flexiblely learn the graph weights over the whole dataset.

\vspace{-3mm}
\paragraph{Hyper-parameter $K$:} As shown in Tab.~\ref{aba-k}, our approach achieves the highest performance when $K$=5. The performance will drop when $K$=3, mainly due to lower proposals recall. When $K$=10, the 
the performance will drop from 58.30\% to 57.15\% because of bringing much noisy proposal candidates. 

\begin{table}[t]
	\centering
	\resizebox{0.25\textwidth}{!}{
		\begin{tabular}{c|c|c|c}
			\hline
			$K$ & 3     & 5     & 10    \\
			\hline
			\textit{Acc\%} & 57.47 & 58.30 & 57.15 \\
	\end{tabular}}
	\caption{\small{Ablation study of $K$ on Flickr30K Entities val set.}}
	\vspace{-3mm}
	\label{aba-k}
\end{table}

\vspace{-2mm}
\subsection{Qualitative Results}
Fig.~\ref{fig:demo} visualizes a variaty of grounding cases of our final results. We can observe that our model is capable of
predicting accurate grounding results when the language description (Fig.~\ref{fig:demo}~\textbf{a}) and visual scene (Fig.~\ref{fig:demo}~\textbf{b}) are complex.
To better understand the capacity of self-taught regression, we also visualize the refined proposals (solid boxes) compared
with their initial proposals (dashed boxes) in Fig.~\ref{fig:demo}~\textbf{c}, and find that object regions can be regressed to a more accurate location, 
e.g., in the upper image the initial proposal of \textit{a red ball} is inaccurate with IoU=0.30, and is refined to a precise region with IoU=0.83.   
In Fig.~\ref{fig:demo}~\textbf{d}, we observe the relation constrain can distinguish the target object from similar candidates, demonstrating 
the effectiveness of such relation-based context information.

\vspace{-3mm}
\section{Conclusion}
In this paper, we propose a flexible context-aware instance representation for weakly
supervised visual grounding by incorporating coarse-to-fine object refinement and entity relation modeling into a two-stage deep network. 
Specifically, we develop a coarse-to-fine denoising strategy, which contains a self-taught regression operation
to refine object proposals and reduce location ambiguities, and adopt a relation constraint
by exploiting language structure to alleviate semantic ambiguities. As a result, we achieve state-of-the-art performance on the public
Flickr30K Entities and ReferItGame benchmarks, outperforming previous work with a sizeable margin.

{\small
\bibliographystyle{ieee_fullname}
\bibliography{egbib}
}

\end{document}


\title{Relation-aware Instance Refinement for Weakly Supervised Visual Grounding\\Supplementary materials}

\author{Yongfei Liu\textsuperscript{\rm 1,5,6}\thanks{Both authors contributed equally. 
This work was done when Yongfei Liu was a research intern at Tencent AI Lab, and Bo Wan was a master student in ShanghaiTech University. This work was supported by Shanghai NSF Grant (No. 18ZR1425100).} 
\quad Bo Wan\textsuperscript{\rm 1,2}\printfnsymbol{1} \quad Lin Ma\textsuperscript{\rm 3} 
\quad Xuming He\textsuperscript{\rm 1,4}\\
\textsuperscript{\rm 1}School of Information Science and Technology, ShanghaiTech University \\
\textsuperscript{\rm 2}Department of Electrical Engineering (ESAT), KU Leuven \\
\textsuperscript{\rm 3}Meituan \quad  \textsuperscript{\rm 4}Shanghai Engineering Research Center of Intelligent Vision and Imaging\\
\textsuperscript{\rm 5}Shanghai Institute of Microsystem and Information Technology,
Chinese Academy of Sciences\\
\textsuperscript{\rm 6}University of Chinese Academy of Sciences\\

\{liuyf3, wanbo, hexm\}@shanghaitech.edu.cn \quad forest.linma@gmail.com}

\maketitle

In this material, we supplement more paper details and  additional deep discussions of several components in our paper as following.



\section{Ranking Loss $\mathcal{L}_{rank}$}
In this section, we depict the ranking loss for image-caption pairs similar to ~[\textcolor{green}{3}].
Specifically, for each image sentence pair $(I, D)$, denote the image representation $ \mathbf{x}_{I}^c $ as the average pooling of the visual features for phrases $\{\mathbf{z}_{i}^c\}_{i=1}^N$ on the coarse-level
and the sentence representation $ \mathbf{x}_{D} $ as the average pooling of the sentence embedding $\mathbf{H}$ in Eq.~\textcolor{red}{1}.
The similarity $S(I,D)$ between $I$ and $D$ in a minibatch $\mathcal{B}$ is defined as the cosine distance of $ \mathbf{x}_{I}^c $ and $ \mathbf{x}_{D}$.
We compute the ranking loss on the coarse-level $\mathcal{L}_{dis}^c$ as follow,
\begin{align}
	\mathcal{L}_{rank}^c = \sum_{D\in\mathcal{B}}\max_{I'\neq I}(0, \Delta - S(I,D) + S(I',D)) \nonumber\\
	+ \sum_{I\in\mathcal{B}}\max_{D'\neq D}(0, \Delta - S(I,D) + S(I,D'))
\end{align}
Similarlly we compute the ranking loss on the fine-level $\mathcal{L}_{rank}^f$ and the total ranking loss $\mathcal{L}_{rank}$ is define as:
\begin{align}
	\mathcal{L}_{rank} = \mathcal{L}_{rank}^c + \mathcal{L}_{rank}^f
\end{align}

\section{Effectiveness of coarse network}
The coarse net aims to select a small set of relevant proposals, which is beneficial to the visual object graph and fine net.
To further investigate the coarse network, we remove it in our model and find the accuracy drops from 58.3\% to 31.2\% dramatically due to severe noise propagation within graph net in this week supervision setting. 

\section{More results for visual object graph network \& relation constraints}
In this paper, we treat the visual object graph network (VOGN) and relation constraints (RC) as a whole, because 
our model is not able to encode visual context cues without VOGN, and cannot restrain the noise 
propagation over VOGN without explicitly relationship supervision.
To investigate the capability of each component, we conduct drop-one-out ablation studies on our final model,
and observe a significant performance drop without any part, as shown in Tab.~\textcolor{red}{1}.


\begin{table}[ht]
	\centering
	\resizebox{0.45\textwidth}{!}{
		\begin{tabular}{cccccc}
			\hline
			Methods  & TSD  & STR               & VOGN&RC             & \textit{Acc\%} \\
			\hline
			ours~(w/o VOGN\&RC)& \checkmark & \checkmark   & -& -                     & 56.88  \\
            ours~(w/o RC)& \checkmark & \checkmark   & \checkmark& -                     & 57.10  \\
            ours~(w/o VOGN)& \checkmark & \checkmark   & -& \checkmark                    & 57.13  \\
			ours     & \checkmark & \checkmark     & \checkmark & \checkmark            & 58.30  \\
            \hline
	\end{tabular}}
	\caption{\small{Ablation Study on Flickr30K Entities val set.}}
\end{table}

\begin{table*}[t]
	\centering
	\resizebox{0.8\textwidth}{!}{
		\begin{tabular}{ccccccccc}
			\hline
			Methods                             & people & clothing & bodyparts & animal & vehicles & instruments & scene & other \\
			\hline
			GroundR[\textcolor{green}{28}] & 44.32  & 9.02     & 0.96      & 46.91  & 46.00    & 19.14       & 28.23 & 16.98 \\
			KAC Net[\textcolor{green}{2}]     & 58.42  & 7.63     & 2.97      & \textbf{77.80} & 69.00    & 20.37       & 45.53 & 17.05 \\
			UTG[\textcolor{green}{39}]       & 58.37  & 14.87    & 2.29      & 68.91  & 55.00    & \textbf{22.22}       & 24.87 & 20.77 \\
			MATN[\textcolor{green}{42}]           & 54.71  & 13.38    & 2.87      & 58.31  & 45.04    & 19.48       & 21.97 & 17.02 \\
			KAC Net*[\textcolor{green}{2}]    & 58.42  & 46.14    & 23.90     & 65.06  & 56.75    & 9.87        & 49.87 & 26.94 \\
			\hline
			ours                                & \textbf{73.70}  & \textbf{54.68}   & \textbf{31.36} & 77.41  & \textbf{69.50}    & 14.81  &\textbf{58.65} & \textbf{37.91} \\
			\hline
	\end{tabular}}
	\caption{\small{Comparison of phrases grounding accuracy over coarse categories on Flickr30K Entities test set.}}
	\label{flickr-coarse-detail}
\end{table*}

\section{Ablation study for four different losses}
Phrase reconstruction loss is a default supervision in this work and has been widely used in weakly-supervised grounding [\textcolor{green}{2,28,19,20}]. 
As shown in Tab.\ref{loss-aba} below, we report the results of using either 
phrase reconstruction loss ($\mathcal{L}_{rec}$) or ranking loss ($\mathcal{L}_{rank}$) in 
the baseline model, and show the results of STR loss ($\mathcal{L}_{reg}$) and RC loss ($\mathcal{L}_{rel}$).  We observe that all the loss terms are effective in model learning.


\begin{table}[h]
	\centering
	\resizebox{0.4\textwidth}{!}{
		\begin{tabular}{cccccc}
			\hline
			Methods  & $\mathcal{L}_{rec}$  & $\mathcal{L}_{rank}$ & $\mathcal{L}_{reg}$~(STR)  &$\mathcal{L}_{rel}$~(VOGN\&RC)& \textit{Acc\%} \\
			\hline
			\multirow{3}{*}{Baseline}& \checkmark  & - & - & - & 47.5  \\
			      & -  & \checkmark & - & - & 43.2  \\
               & \checkmark  & \checkmark & -           & -                   & 48.18  \\
         \hline
         \multirow{3}{*}{Baseline + TSD}& \checkmark  & \checkmark & -           & -                   & 50.80  \\
			         &  \checkmark & \checkmark & \checkmark  & -                   & 56.88  \\
                  & \checkmark  & \checkmark & \checkmark  & \checkmark          & 58.30  \\

			\hline
	\end{tabular}}
	\caption{\small{Ablation Studies of four losses on Flickr30K val set.}}
	\label{loss-aba}
\end{table}

\section{Relation types encoded in graph network}

Our graph net mainly captures semantic and spatial relations, and encodes spatial cues (feature locations) in object feature as in [\textcolor{green}{21}].
Concretely, we select top-88 frequent relation types on Flickr30K (63\% for semantic and 37\% for spatial) and top-34 relations on ReferitGame (34\% for semantic, 66\% for spatial).

We further investigate the efficacy of relation encoding in Flickr30K,
and report relations classification accuracy in Tab.~\ref{rc_res} below. It shows our relation encoding module can capture semantic and spatial relations indeed.

\begin{table}[h]
	\centering
	\resizebox{0.4\textwidth}{!}{
		\begin{tabular}{ccccc}
			\hline
         &  \# classes & top-1~(\%) & top-5~(\%) & top-10~(\%)  \\
			\hline
			semantic  & 67  & 41.2  & 79.1 & 88.6  \\
			spatial  & 21 & 53.1  & 84.6 & 91.8  \\
		 	all      & 88 & 45.6  & 81.1 & 89.8    \\ 
			\hline
	\end{tabular}}
	\caption{\small{Relation classification results on Flickr30K val set.}}
	\label{rc_res}
\end{table}

\section{Coarse Categories Accuracy}
As shown in Tab.~\ref{flickr-coarse-detail}, our method outperforms the previous state-of-the-art in most
coarse categories in Flickr30k test set, which validates the effectiveness of our network. In addition, our model
performs inferior result in instruments category, which is caused by lower proposal recall when using object detector pretrained on Visual Genome dataset.
We find that most instruments phrases are class "guitar", which is not contained in Visual Genome class space.  

\section{Comparison with Concurrent Work}
We compare our model with concurrent work MAF Net~[\textcolor{green}{35}], of which feature extractor
is pretrained with additional supervision from object attributes on Visual Genome dataset.
For a fair comparison and keeping in line with the previous works~[\textcolor{green}{2, 7}],
we re-implement their released code with the same 
feature maps as ours, denoted as MAF*.
As shown in Tab.~\textcolor{red}{2}, we outperforms MAF Net with 1.01\% grounding accuracy, which
validates the superiority of our proposed flexible and context-aware object representation for weakly supervised visual grounding.
\begin{table}[ht]
	\centering
	\resizebox{0.2\textwidth}{!}{
		\begin{tabular}{cc}
			\hline
			Methods  & \textit{Acc\%} \\
			\hline
			MAF*~[\textcolor{green}{35}]& 58.26  \\
			ours  & 59.27  \\
            \hline
	\end{tabular}}
	\caption{\small{Comparison with concurrent work on Flickr30K Entities test set.}}
\end{table}

\section{Implementation details for ReferItGame dataset}
For the visual feature extraction, we take the same object detector pretrained on Visual Genome to generate $M$=50 object proposals and compute their visual representation
via RoI-Align. We also select $K$=5 proposals in coarse-level matching network to suppress most of the background distractors. For the semantic relations, we select top $C_r$=34 relations 
whose frequency are greater than 10. It worth noting that we explicitly parse the expression in ReferItGame dataset into $\langle$subject, relation, object$\rangle$ pairs following KPRN~[\textcolor{green}{20}],
and regard the subject as target grounding phrase.

For model learning, we keep the same training configuration as in Flickr30k Entities but the initial learning rate is set as 0.005.